\documentclass[runningheads]{llncs}

\usepackage{cite}
\usepackage{amsmath,amssymb,amsfonts}
\usepackage{algorithmic}
\usepackage{graphicx}
\usepackage{textcomp}
\usepackage{xcolor}
\usepackage{multirow}
\usepackage{booktabs}
\usepackage{caption}
\usepackage{array}
\usepackage{acronym}
\usepackage{rotating} 
\usepackage{geometry}
\usepackage{placeins}
\usepackage{hyperref}
\usepackage{amssymb}

\begin{document}
\setlength{\textfloatsep}{10pt plus 2pt minus 2pt}
\setlength{\floatsep}{8pt plus 2pt minus 2pt}
\setlength{\intextsep}{8pt plus 2pt minus 2pt}

\title{Adaptive Confidence-weighted Expansion for Trustworthy Multi-Omics Multimodal Fusion}
\titlerunning{Adaptive Confidence-weighted Expansion for Trustworthy Multi-Omics Multimodal Fusion}
%
%
\author{Mohammad Raahemi\inst{1}\orcidID{0000-0002-3229-570X} \and
Ali Sekhavati\inst{1}\orcidID{0000-0002-4554-2749} \\
Alireza Maleki\inst{2}\orcidID{0009-0007-8550-8194} \and
Hamid Nasiri\inst{3,*}
\orcidID{0000-0002-9279-6063}}
\authorrunning{M. Raahemi et al.}
%
\institute{Faculty of Engineering, University of Ottawa, Ottawa ON K1N 1A2, Canada \and
RIV Lab, Department of Computer Engineering, Bu-Ali Sina University, Hamedan, Iran
\and
School of Computing and Communications, Lancaster University, Lancaster, UK \\
\email{mrahe103@uottawa.ca}
\\
\email{ali.sekhavati@uottawa.ca}
\\
\email{alireza.maleki@eng.basu.ac.ir}
\\
\email{h.nasiri@hud.ac.uk}}
\maketitle              
\begin{abstract}
Multimodal learning is a robust approach to improve predictive performance in applications such as medical prognosis. However, the clinical applicability of models that use multimodal learning is hampered by their poor performance under noisy or uninformative data streams. Present fusion approaches often lack robust mechanisms for the dynamic assessment of data quality and for the provision of a trustable confidence score on the final prediction. This dissuades their deployment in safety-critical settings. To address these limitations, we introduce Adaptive Confidence-weighted Expansion (ACE), a novel framework to enhance the trustworthiness of multimodal fusion models. ACE first enhances the multimodal space by generating new, complementary modalities from intra-modality correlations. It then employs a dual-level confidence mechanism that (1) adaptively reweighs all modalities by their reliability before fusion and (2) estimates a global trust score over the fused, final decision. To evaluate ACE, we used four challenging multi-omics datasets (BRCA, KIPAN, LGG, and ROSMAP). ACE significantly outperforms existing state-of-the-art algorithms in both classification performance and confidence calibration. Our framework provides a more stable and robust data fusion method that facilitates the use of multimodal learning in addressing high-stakes problems. The source code is publicly available at: \href{https://github.com/AliGreen0/ACE-Dynamics}{https://github.com/AliGreen0/ACE-Dynamics} 

\end{abstract}
\section{Introduction}
Multimodal learning integrates heterogeneous data sources, which enables models to leverage complementary information between modalities and boost predictive performance \cite{bioengineering11030219}. Multimodal approaches integrate different modalities into a unified representation using neural networks \cite{hu2021unit}. Many studies have integrated heterogeneous data sources, including imaging, genomic, and clinical data —to improve predictive accuracy. For instance, Huang et al. \cite{Huang2022MedImageAnal} combined radiomic and genomic features for glioma grading, achieving a 6\% improvement in AUC compared to prior studies, while Chaudhary et al. \cite{Chaudhary2018NatCommun} integrated transcriptomic and clinical data for liver cancer survival prediction, demonstrating the potential of multimodal learning in biomedical prognosis. However, current multimodal learning tools face limitations \cite{wang2021survey}. Existing fusion strategies struggle to assess data accuracy and to produce confidence scores for final predictions. As a result, the deployment of multimodal models is limited in safety-critical scenarios such as computer-aided diagnosis. More reliable fusion tools are needed to safely leverage multimodal learning in these settings. 

Several barriers exist to the use of multimodal fusion in medical contexts, including a lack of dynamic assessment tools, a focus on image-based analysis, and the quality of the training data. Traditional multimodal fusion methods aim to obtain a joint representation by concatenating features or designing specific neural architectures. Early fusion integrates modalities directly at the data level \cite{poria2015deep}, intermediate fusion combines modality-specific features through dedicated network layers \cite{hong2020more, arevalo2017gated}, and decision-level fusion aggregates outputs based on predictive uncertainty \cite{han2022trusted, natarajan2012multimodal}. While effective, these approaches struggle to dynamically evaluate the informativeness of features and modalities across different samples, which is essential for trustworthy multimodal classification  \cite{han2022multimodal}. Additionally, multimodal learning is typically deployed to assess both images and clinical data.  However, little research exists on the use of multimodal learning that does not incorporate imaging. Finally, in medical datasets, uninformative features and noisy modalities are common \cite{argelaguet2021statistical, yelipe2018efficient}. These challenges further highlight the need for robust fusion strategies.

To improve fusion stability and explainability, Han et al. \cite{han2022multimodal} proposed Multimodal Dynamics (MM-Dynamics), which jointly models feature-level and modality-level informativeness. Their approach introduces sparse gating to dynamically select informative features and employs true-class probability to capture modality confidence. Extensive experiments on multiple medical classification datasets demonstrated the superiority of MM-Dynamics compared with state-of-the-art methods \cite{wang2021mogonet, han2022trusted, hong2020more , arevalo2017gated}.
Although MM-Dynamics represents a significant advancement in multimodal learning, several challenges remain. First, the framework directly relies on the original set of modalities without exploring intra-modality correlations that could enrich representations. Second, while modality confidence is modeled, the framework does not adaptively reweight modality contributions, limiting robustness under uncertain conditions. Third, although the method estimates modality-aware confidence, it does not introduce model-level (global) predictive certainty, leaving end-to-end reliability under-specified . In this paper, we propose Adaptive Confidence-weighted Expansion (ACE), a novel multimodal fusion framework that extends and improves MM-Dynamics  by addressing these limitations. Our contributions are as follows:

\begin{enumerate}
    \item \textbf{Correlation-based modality expansion:} For datasets with \textit{n} original modalities (e.g., BRCA \cite{cancer2012comprehensive} and ROSMAP \cite{a2012overview}), we generate additional derived modalities by computing intra-modality correlations and applying dot-product transformations with their correlated versions. This process yields \textit{2n} modalities derived from the original \textit{n}, thereby enriching the multimodal feature space and enhancing the overall representation capacity.
    \item \textbf{Confidence-weighted fusion:}  At the final fusion stage, where per-view confidence loss is calculated, we introduce an adaptive weighting mechanism. Low-confidence modalities are penalized, while reliable modalities are emphasized, leading to more trustworthy fusion.
    \item \textbf {Global TCP for fused trust estimation:} ACE adds a single global TCP head on the post-fusion representation with the goal of estimating the confidence of the final decision. Trained to predict the fused models true-class probability, this head provides a calibrated sample-level trust signal that captures cross-modal agreement/conflict after fusion,  and stabilizes training—especially when individual modalities are noisy or partially missing.
    \item \textbf {Comprehensive evaluation:} We conduct experiments on BRCA \cite{cancer2012comprehensive}, ROSMAP \cite{a2012overview}, KIPAN \cite{wang2021mogonet} and LGG \cite{buda2019association} datasets, showing that ACE consistently outperforms MM-Dynamics and other state-of-the-art methods in terms of accuracy, stability, and reliability.
    
\end{enumerate}

\section{Literature Review}
The increased application of deep learning (DL) methods in healthcare has ushered in new possibilities for clinical prognostication and disease classification. Traditionally, prognostication using clinical data involves statistical models, which often struggle to efficiently capture high-dimensional non-linear interactions. With the advent of DL architectures, researchers can leverage vast amounts of clinical information to predict disease progression with greater accuracy. 

Many DL models for diagnosis and prognosis use medical imaging data, and the use of DL to evaluate images has shown promising results in diagnosing illnesses such as breast cancer.\cite{maleki2023breast} However, deep learning on clinical (non-image) data is increasingly outperforming traditional statistical models of prognosis. Although multimodal fusion (imaging + genomics + clinical) generally boosts accuracy, a focused body of evidence  shows that even unimodal clinical deep models can surpass conventional methods; an important practical consideration when imaging is unavailable or cost-prohibitive.

At the same time, the literature exhibits notable gaps. Much of the available evidence concentrates on multimodal pipelines. Leaving relatively few studies that evaluate ANN, CNN, and RNN architectures using strictly clinical features. These gaps in the existing literature point toward a need for clearer benchmarks: curation and standardization of high-quality, clinical datasets; and transparent reporting of preprocessing, hyperparameters, and training setups to ensure reproducibility. There is a  need for methods that exploit clinical signals effectively while remaining robust, interpretable, and evaluation-ready.
Recent advancements in precision oncology have emphasized the critical role of classifying cancer into molecular subtypes, as this classification significantly influences early diagnosis and the development of targeted therapies \cite{mateo2022delivering}. Accurate identification of these subtypes is essential, and numerous methodologies have been proposed over the past decades to leverage multi-omics data for cancer classification tasks \cite{li2021cancer}.

Data-driven approaches have shown substantial advantages across various domains by utilizing the robust feature extraction capabilities of deep neural networks \cite{wang2021single,nicora2020integrated,lecun2015deep}. In the context of multi-omics data integration, existing models predominantly adopt either early fusion or late fusion strategies. Early fusion methods concatenate multi-omics data before feeding them into a deep neural network for feature extraction. In contrast, late fusion approaches involve independent feature extraction from each omic type, followed by the integration of these features for final classification. Given that multiomics  data can be treated as multi-modal data, various multi-modal learning models have been introduced for their interpretation, often outperforming single-omics approaches \cite{wang2021mogonet,chan2022combining}. A fundamental challenge in multi-modal learning is effectively fusing complementary information from different modalities.

Feature-level fusion techniques integrate raw data or extracted features through operations such as concatenation or summation \cite{dalla2015challenges}. However, these methods may struggle with heterogeneous or inconsistently scaled data. Decision-level fusion strategies, on the other hand, combine predicted labels from each modality to reach a final decision  \cite{wang2021mogonet}. More recently, multi-level fusion approaches have been proposed, which combine both feature- and decision-level information to capture low-level and high-level modality characteristics simultaneously  \cite{hu2021unit}.

With increasing awareness of data and modality uncertainty, the focus has shifted toward trustworthy multi-modal learning. To address this, Han et al. \cite{han2022multimodal} introduced a dynamic fusion network that incorporates sparse gating to capture within-modality variation and leverages class probability to evaluate modality-wise classification confidence. Similarly, Wang et al. proposed MOGONET, a novel multi-omics integration framework that employs a View Correlation Discovery Network (VCDN) to explore both omics-specific and cross-omics correlations for more effective biomedical classification \cite{wang2021mogonet,wang2019generative}. Rather than fusing low-level features, MOGONET utilizes VCDN to discover higher-level inter- and intra-modality relationships in the label space to achieve robust classification outcomes.

Further extending the notion of trustworthy learning, Ma et al. \cite{ma2021trustworthy} proposed the Mixture of Normal-Inverse Gamma (MoNIG) model, which estimates uncertainty in modality data to guide adaptive integration and yield reliable regression results. Han et al. \cite{han2022multimodal,han2022trusted}, building on Dempster–Shafer theory, introduced a variational Dirichlet framework that characterizes class probability distributions and integrates different modalities at the evidence level, thus ensuring both robustness and reliability against noisy or incomplete data.

Graph-based learning approaches have also been increasingly adopted to exploit the structural properties inherent in omics data. Unlike traditional Euclidean-based methods, Graph Convolutional Networks (GCNs) operate on graph-structured data, where each node represents a sample and edges capture meaningful relationships. Ramirez et al. incorporated intra-omic, protein–protein interaction, and gene co-expression networks into a GCN framework for enhanced feature aggregation  \cite{ramirez2020classification}. Nevertheless, standard GCNs often capture only local neighborhood information. To address this limitation, Li et al. introduced a parallel GCN model that integrates gene-based prior knowledge graphs for cancer subtype classification \cite{li2021cancer}. Additionally, Wang et al. developed a GCNbased model that constructs a cell–cell similarity graph using K-Nearest Neighbors (KNN) to classify single-cell sequencing data \cite{wang2021single}.

Despite significant progress, several challenges persist in multi-omics data classification. The high rate of missing values in multi-omics datasets can distort final feature representations if not properly handled. Also, most deep multimodal learning frameworks rely on simple modality-wise fusion strategies which may not fully exploit the intricate cross-modal relationships within samples. Finally, contrastive learning-based methods often focus solely on aligning sample-level positive and negative pairs while overlooking the underlying cluster structures among different samples.

To address these challenges, we propose a novel framework, ACE, which integrates feature informativeness, modality certainty, and cross-modal sample relationships to produce a more representative and consensus-based embedding for final classification.

\section{Methodology}

\subsection{Network Architecture}

\begin{figure*}[!t]
    \centering
    \includegraphics[width=\textwidth]{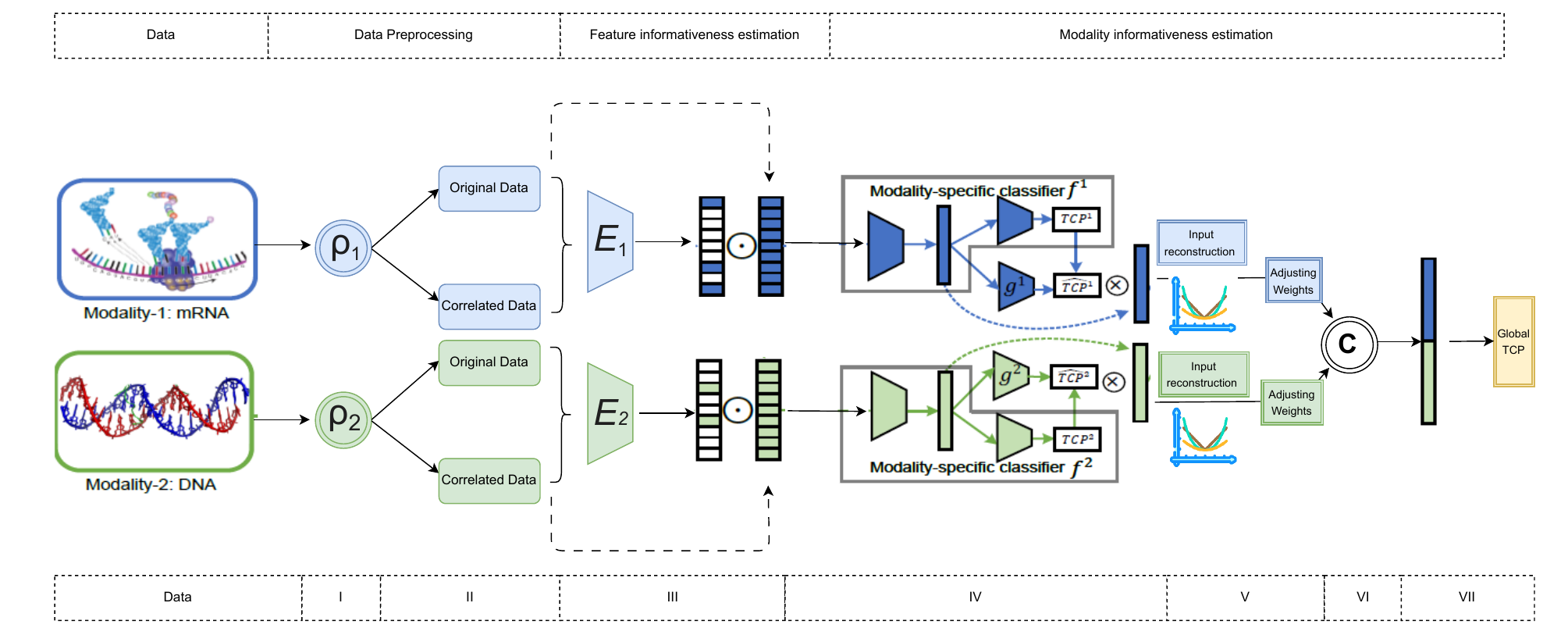}
    \vspace{10pt}
    \caption{Framework of ACE multimodal dynamics. We use a two-modality case for better illustration. The proposed method is mainly
composed of the following steps. I. Calculation of correlation, II. Adding the generated views  to data, III. Feature Informativeness Estimation, IV. Modality specific classifier, V. Adjusting weighting including the Loss Function and Input Reconstruction, VI. Concatenating Modalities, VII. Global TCP}
    \label{fig:architecture}
\end{figure*}

The proposed ACE framework extends the MM-Dynamics architecture \cite{han2022multimodal} by introducing correlation-based modality expansion, confidence-weighted fusion, a global confidence estimation module,  and an input reconstruction layer. The architecture consists of four major components: (1) feature encoding with feature-level gating, (2) modality-specific classification and confidence estimation, (3) confidence-weighted fusion, and (4) global TCP. An overview is illustrated in Figure \ref{fig:architecture}.

\subsubsection{Summary of Architecture:}

The datasets considered in this study include BRCA (breast cancer), LGG (lowgrade glioma), KIPAN (kidney cancers), and ROSMAP, which comprises longitudinal clinical and cognitive data related to aging and Alzheimer’s disease. To perform classification within BRCA, ROSMAP, KIPAN, and LGG datasets, we adopted a multi-view representation learning approach. For each modality, we first constructed an additional view by computing the Pearson correlation matrix of the features and multiplying it with the corresponding input representation. This step enriched the data by emphasizing inter-feature relationships and provided complementary structural information before model training. To mitigate the issue of insufficient data within the different datasets, we introduced noise and performed input reconstruction on each dataset. The resulting expanded set of views was then fed into a multi-view neural architecture.

In ACE, each view passes through a dedicated encoder that applies feature-level attention, nonlinear transformations, and dropout-based corruption to encourage robust feature learning. Each view is paired with a TCP module, which simultaneously produces class logits and a confidence estimate. These confidence scores serve two purposes: (i) they are used for adaptive weighting, reducing the influence of unreliable views in the loss function, and (ii) they guide feature scaling during fusion. The encoded features from all views are concatenated to form a joint representation, which is fed into a shared classifier for the final prediction. Importantly, this fused representation is also passed through a global TCP layer, providing a data-level confidence estimate for the overall prediction.

\subsubsection {Correlation:}

In high-dimensional biological datasets such as gene expression or DNA methylation, features often exhibit complex co-expression patterns . Encoding this structure can help the model better understand intra-modality dependencies. To capture the internal structure of each modality and enhance the representation of individual features, we applied a correlation-based transformation to the input data. For each modality $X_i$, we computed the pairwise Pearson correlation matrix $C_i$ using the training samples. This matrix reflects the degree of linear association between features within the same modality.

Let \( X_i \in \mathbb{R}^{n \times d} \) denote the training data matrix for modality \( i \), where \( n \) is the number of samples and \( d \) is the number of features. We transformed the original features by computing the dot product of the input features with their corresponding correlation matrix:

\begin{equation}
X_i' = X_i \cdot corr(X_i) ,
\end{equation}

This transformation can be interpreted as a projection of each sample into a space where each feature is re-expressed as a weighted combination of all other features within the same modality, with weights reflecting their pairwise correlations. The objective is to emphasize consistently co-expressed features and reduce the influence of noisy or weakly informative dimensions.

This operation was applied to both the training and test sets. The correlation matrix \( C_i \) was computed exclusively from the training data to avoid information leakage.

\begin{figure*}[!t]
    \centering
    \includegraphics[width= 0.65\textwidth]{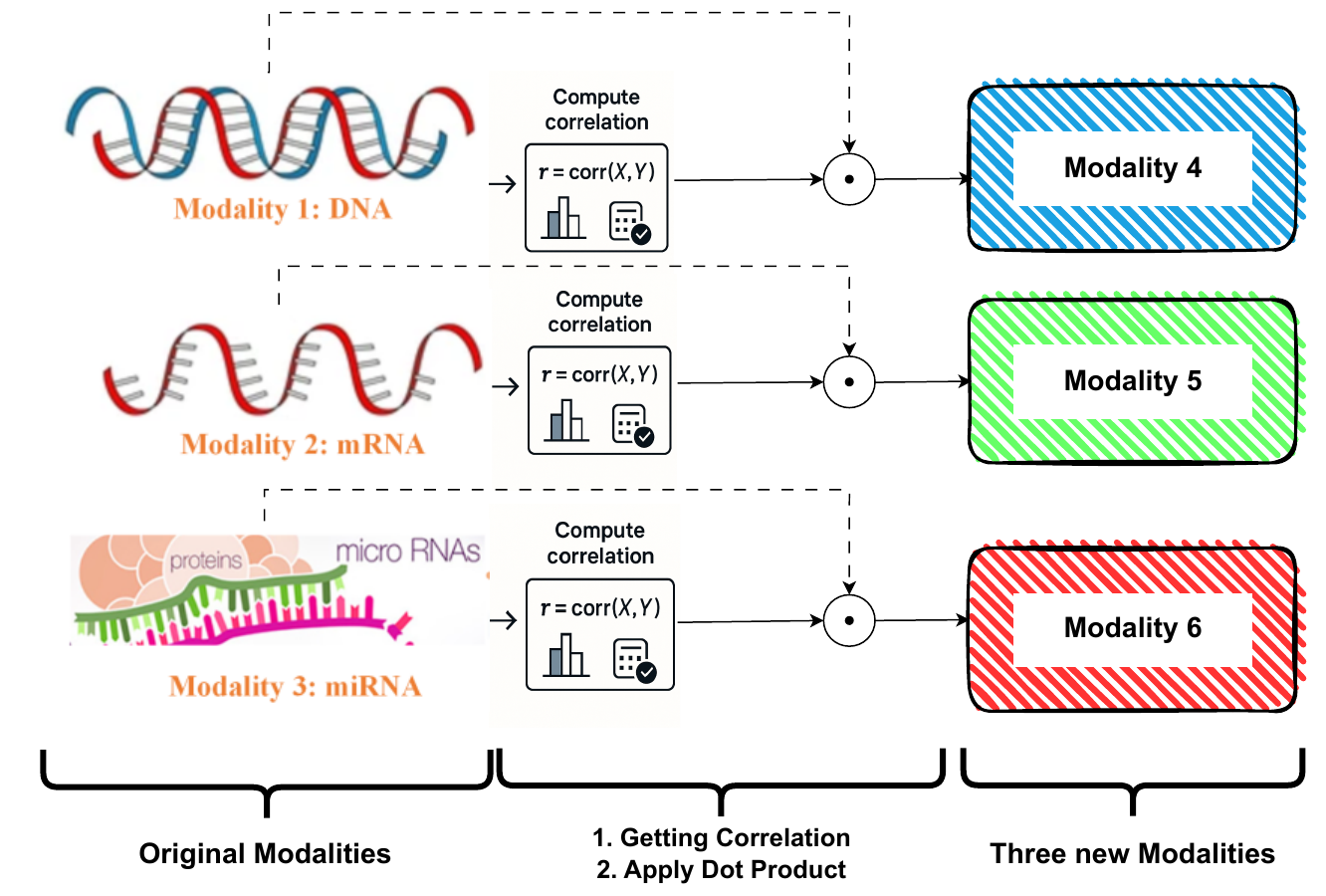}
    \vspace{10pt}
    \caption{Illustration of Correlation-driven Modality Augmentation.}
    \label{fig:correlation}
\end{figure*}

Figure \ref{fig:correlation} illustrates the correlation-based transformation process. As shown in the figure, the pipeline consists of two main steps: (1) calculating the correlation matrix from the training data, and (2) applying the dot product to generate the transformed modalities. This visual representation highlights how the transformation encodes intra-modality dependencies before feeding the data into subsequent modeling stages.

\subsubsection{Noise-Invariant Gating and Encoding:}

To enhance robustness against noise and missing values, each modality input $X_i$ is deliberately corrupted during training using dropout and additive Gaussian perturbations:
\begin{equation}
\tilde{X}_i = \text{Dropout}(X_i, p) + \epsilon, \quad \epsilon \sim \mathcal{N}(0, \alpha^2) ,
\end{equation}
where:
\begin{equation}
\hat{X} = \left[ X_1, X_2, \dots, X_M, X_1', X_2', \dots, X_M' \right] ,
\end{equation}
This corruption step encourages the model to learn stable, noise-invariant representations.  

Next, a feature-level gating mechanism estimates the relative importance of each feature dimension through a learned sigmoid transformation:
\begin{equation}
G_i = \sigma(W_i \tilde{X}_i), \quad Z_i = \tilde{X}_i \odot G_i,
\end{equation}
where $W_i$ is a trainable projection, $\sigma(\cdot)$ denotes the sigmoid function, and $\odot$ the elementwise product. This operation suppresses uninformative or noisy features while preserving salient ones.  

The gated features $Z_i$ are then passed through a modality-specific encoder network $f_i(\cdot)$:
\begin{equation}
h_i = f_i(Z_i), \quad h_i \in \mathbb{R}^{n \times d_h},
\end{equation}
implemented as a fully connected layer with batch normalization, ReLU activations, and dropout.  

The resulting latent representations serve as the basis for downstream tasks, including classification, confidence estimation, and reconstruction (detailed in the following subsection). This design ensures that the encoder outputs are both noise-robust and information-preserving across modalities.

\subsubsection{Input Reconstruction as Self-Supervised Regularization:}

Deep neural networks trained on limited or imbalanced biological datasets are prone to overfitting. To mitigate this issue, we drew inspiration from recent advances in self-supervised learning, where recovering masked or corrupted inputs has been shown to improve generalization.  

In our framework, the corrupted modality inputs $\tilde{X}_i$ (described in the previous subsection) are passed through a modality-specific encoder–decoder pathway. The encoder produces a latent representation $h_i = f_{\phi}(\tilde{X}_i)$, and the decoder attempts to reconstruct the original input $X^{(r)}_i = g_{\theta}(h_i)$. The reconstruction objective is defined as:
\begin{equation}
\mathcal{L}_{\text{rec}, i} = \lambda_{\text{rec}} \cdot \| X_i - X^{(r)}_i \|_2^2 ,
\end{equation}

This reconstruction loss, combined with the classification and confidence calibration objectives, acts as a self-supervised regularizer. It compels the encoder to retain modality-specific structure under corruption, thereby reducing sensitivity to noise or missing values and improving the stability and generalization of multimodal representations.

\subsubsection{Adaptive weighting:}

In multimodal learning, different modalities often exhibit varying levels of reliability due to noise, missing values, or modality-specific biases. A naive aggregation that treats all modalities equally may therefore propagate uncertainty from weaker modalities, ultimately degrading predictive performance. To address this challenge, we introduce an adaptive weighting scheme that dynamically adjusts each modality's contribution to the loss function based on its confidence.  

For each modality $i$, let $z_i$ denote the classification logits, and let $\alpha_i \in \mathbb{R}$ represent its scalar confidence score predicted by a modality-specific confidence layer. The per-modality classification loss is then reweighted according to its confidence:
\begin{equation}
\mathcal{L}_{\text{aw}, i} =\lambda_{\text{aw}} \cdot  \sigma (\alpha_i) \cdot \text{CE}(z_i, y),
\end{equation}
where $\text{CE}(\cdot)$ shows the cross-entropy loss and $\sigma (\alpha_i)$ represents a normalized weight for modality $i$.

The overall adaptive loss across all modalities is defined as:
\begin{equation}
\mathcal{L}_{\text{adaptive}} = \frac{1}{m} \sum_{i=1}^{m} \mathcal{L}_{\text{aw}, i},
\end{equation}
with $m$ denoting the number of modalities.  

This formulation ensures that modalities with higher predicted confidence contribute more strongly to the optimization, while uncertain modalities are down-weighted. By allowing the model to learn these weights dynamically during training, the framework achieves a better balance between leveraging strong modalities and mitigating the effect of noisy or unreliable ones.

\subsubsection{Modality-Specific TCP:}
Following MM-Dynamics \cite{han2022multimodal}, each encoded modality representation $h_i$ is passed through a modality-specific Trustworthy Calibration Prediction (TCP) module. This module consists of two parallel heads: a classification head and a confidence head.

The classification head is a fully connected layer that maps the encoded features to class logits:
\begin{equation}
z_i = W_i^{(\text{cls})} h_i + b_i^{(\text{cls})}, \quad 
p_i = \text{softmax}(z_i), \quad p_i \in \mathbb{R}^{n \times c},
\end{equation}
where $c$ is the number of classes and $p_i$ denotes the calibrated prediction for modality $M_i$.

The confidence head is a separate linear layer which outputs a scalar confidence score for each sample:
\begin{equation}
\alpha_i = \sigma\!\left(W_i^{(\text{conf})} h_i + b_i^{(\text{conf})}\right), 
\quad \alpha_i \in [0,1]^{n \times 1} ,
\end{equation}
where $\alpha_i$ reflects the reliability of the modality’s prediction.

After that, the encoded modality representation is then reweighted by its confidence before multimodal fusion:
\begin{equation}
\hat{h}_i = \alpha_i \odot h_i ,
\end{equation}

Thus, each modality contributes both a calibrated class prediction $p_i$ and a confidence-weighted feature $\hat{h}_i$ to the subsequent fusion stage.

\subsubsection{Global TCP:}

The confidence-weighted representations from all modalities, $\{\hat{h}_i\}_{i=1}^m$, are concatenated into a joint feature vector $H$, which is processed by a shared multilayer perceptron to obtain a fused embedding $u$. This embedding serves as the basis for the final prediction.

To explicitly model uncertainty after fusion, we introduce a \emph{Global TCP} module. Unlike modality-specific TCPs, which provide per-modality predictions and confidence estimates, the global TCP operates on the fused embedding $u$ and produces:

\begin{multline}
p_{\text{global}} = \text{softmax}\!\left(W^{(\text{cls})} u + b^{(\text{cls})}\right),
\alpha_{\text{global}} = \sigma\!\left(W^{(\text{conf})} u + b^{(\text{conf})}\right)
\end{multline}

where $p_{global} \in \mathbb{R}^{n \times c}$ is the final multimodal prediction and $\alpha_{global} \in [0,1]^{n \times 1}$ is the associated global confidence score.

This design provides two complementary benefits:  
(1) modality-specific TCPs ensure that each input stream contributes with calibrated reliability, and  
(2) the global TCP captures residual uncertainty that emerges only after cross-modal fusion. Together, these mechanisms enable both fine-grained (per-modality) and holistic (global) confidence estimation, which is critical for trustworthy multimodal decision-making.

\subsubsection{Loss Function:}

The total training loss integrates multimodal classification, modality-specific classification and confidence calibration (TCP), input reconstruction, adaptive weighting and feature sparsity.

Let $p_{global}$ denote the global logits and $\alpha_{global}$ the global confidence score. The global classification loss is:
\begin{equation}
\mathcal{L}_{\text{cls}} = \frac{1}{n} \sum_{j=1}^{n} \text{CE}\left( p_{\text{global}_j}, y_j \right),
\end{equation}

where $n$ is the minibatch size. The global TCP loss is
\begin{equation}
\mathcal{L}_{\text{tcp-global}} = \lambda_{\text{tcp-g}} \cdot \frac{1}{n} \sum_{j=1}^{n} \big\| \alpha_{\text{global}_j} - p_{\text{pred}_j} \big\|_2^2,
\end{equation}
where $p_{\text{pred}_j}$ is the predicted probability of the correct class. The combined global loss is
\begin{equation}
\mathcal{L}_{\text{global}} = \mathcal{L}_{\text{cls}} + \mathcal{L}_{\text{tcp-global}} ,
\end{equation}

For each modality $i$, with logits $z^{(i)}$ and confidence $c^{(i)}$, the classification and calibration loss is
\begin{equation}
\mathcal{L}_{\text{tcp},i} = \lambda_{\text{tcp}} \cdot \Big( \text{CE}(z^{(i)}, y) + \| c^{(i)} - p_{\text{pred}} \|_2^2 \Big),
\end{equation}
 
To encourage compact representations, the mean of the gating outputs $G_i$ is penalized:
\begin{equation}
\mathcal{L}_{\text{feat},i} = \lambda_{\text{feat}} \cdot \frac{1}{d} \sum_{k=1}^{d} G_{i,k} ,
\end{equation}
where $k$ indicates the feature dimensions of modality $i$. Finally, the full training objective is
\begin{equation}
\mathcal{L}_{\text{total}} = \mathcal{L}_{\text{global}} + \sum_{i=1}^{m} \Big( \mathcal{L}_{\text{tcp},i} + \mathcal{L}_{\text{rec}, i} + \mathcal{L}_{\text{aw}, i} + \mathcal{L}_{\text{feat},i} \Big) ,
\end{equation}


\section{Experiments and Results}

In this section, we compare the proposed ACE model with several state-of-the-art classification methods on four real-world multimodal datasets. Extensive experimental results demonstrate the superior performance of our approach relative to existing methods. Additionally, we conduct ablation studies to assess the contribution of individual components within the proposed network.
All experiments were conducted on a Linux workstation running Ubuntu 20.04.1, equipped with four NVIDIA GeForce RTX 3080 Ti and an Intel Xeon Gold 6254 CPU (3.10 GHz).

We compared five representative multimodal learning methods—MOGONET \cite{wang2021mogonet}, TMC \cite{han2022trusted}, CF \cite{hong2020more}, GMU \cite{arevalo2017gated}, and Dynamics\cite{han2022multimodal}. MOGONET leverages a Graph Convolutional Network together with a View Correlation Discovery Network to model inter-omics dependencies for classification. TMC performs decision-level fusion by weighting modality-specific predictors according to their estimated confidence. CF implements late fusion by concatenating high-level, modality-specific representations prior to the final classifier. GMU constructs an intermediate joint representation that adaptively combines modalities (via a gated multimodal unit), enabling cross-modal information flow. Finally, MM-Dynamics adopts a dynamic fusion scheme that evaluates both feature-level and modality-level relevance to integrate multimodal evidence.
To evaluate the performance of the compared methods, we employ different metrics depending on the classification task. For binary classification datasets, we report accuracy (ACC), $F_1$-score ($F_1$), and the area under the ROC curve (AUC). For multi-class classification datasets, we use accuracy (ACC), weighted $F_1$-score ($F_1$ weighted), and macro-averaged $F_1$-score ($F_1$ macro). Following prior work \cite{wang2021mogonet,han2022multimodal}, each dataset is randomly split into training and testing sets, and experiments are repeated 20 times to mitigate the effects of partition bias. We report the average performance and standard deviation across these runs. Network training is performed using the Adam optimizer with learning rate decay. Each experiment is run for 1200 epochs, after which testing results are recorded. The comparison results are shown in Table ~\ref{tab:comparison}.


\begin{table*}[t!]
\centering
\caption{Comparison with state-of-the-art methods on multi-omics classification. \textbf{Bold} value indicates the best performance.}
\label{tab:comparison}
\resizebox{\textwidth}{!}{%
\begin{tabular}{l c c c c c c c c c c c c}
\toprule
\multirow{3}{*}{\textbf{Method}} & \multicolumn{3}{c}{\textbf{BRCA}} & \multicolumn{3}{c}{\textbf{KIPAN}} & \multicolumn{3}{c}{\textbf{LGG}} & \multicolumn{3}{c}{\textbf{ROSMAP}} \\
\cmidrule(lr){2-4} \cmidrule(lr){5-7} \cmidrule(lr){8-10} \cmidrule(lr){11-13}
& ACC & Weighted $F_1$ & Macro $F_1$ & ACC & Weighted $F_1$ & Macro $F_1$ & ACC & $F_1$ & AUC & ACC & $F_1$ & AUC \\
\midrule
KNN & 74.2$\pm$2.4 & 73.0$\pm$2.5 & 68.2$\pm$2.5 & 96.7$\pm$1.1 & 96.7$\pm$1.1 & 96.0$\pm$1.4 & 72.9$\pm$3.4 & 73.8$\pm$3.8 & 79.9$\pm$3.8 & 65.7$\pm$3.6 & 67.1$\pm$4.5 & 70.9$\pm$4.5 \\
SVM & 72.9$\pm$1.8 & 70.2$\pm$1.7 & 64.0$\pm$1.7 & 99.5$\pm$0.3 & 99.5$\pm$0.3 & 99.4$\pm$0.4 & 75.4$\pm$4.6 & 75.7$\pm$4.6 & 75.4$\pm$4.6 & 77.0$\pm$2.4 & 77.8$\pm$2.6 & 77.0$\pm$2.6 \\
LR  & 73.2$\pm$1.2 & 69.8$\pm$2.6 & 64.2$\pm$2.6 & 97.4$\pm$0.2 & 97.4$\pm$0.2 & 97.2$\pm$0.4 & 76.1$\pm$1.8 & 76.7$\pm$2.7 & 82.3$\pm$2.7 & 69.4$\pm$3.7 & 73.0$\pm$3.5 & 77.0$\pm$3.5 \\
RF  & 75.4$\pm$0.9 & 73.3$\pm$1.3 & 64.9$\pm$1.3 & 98.1$\pm$0.6 & 98.1$\pm$0.6 & 97.5$\pm$1.1 & 74.8$\pm$1.2 & 74.2$\pm$1.0 & 82.3$\pm$1.0 & 72.6$\pm$2.9 & 73.4$\pm$1.9 & 81.1$\pm$1.9 \\
NN  & 75.4$\pm$2.8 & 74.0$\pm$4.7 & 66.8$\pm$4.7 & 99.1$\pm$0.5 & 99.1$\pm$0.5 & 99.1$\pm$0.5 & 73.7$\pm$2.3 & 74.8$\pm$3.7 & 81.0$\pm$3.7 & 75.5$\pm$2.1 & 76.4$\pm$2.5 & 82.7$\pm$2.5 \\
\midrule
GRridge \cite{van2016better} & 74.5$\pm$1.6 & 72.6$\pm$2.5 & 65.6$\pm$2.5 & 99.4$\pm$0.4 & 99.4$\pm$0.4 & 99.3$\pm$0.4 & 74.6$\pm$3.8 & 75.6$\pm$4.4 & 82.6$\pm$4.4 & 76.0$\pm$3.4 & 76.9$\pm$2.3 & 84.1$\pm$2.3 \\
BPLSDA \cite{singh2019diablo} & 64.2$\pm$0.9 & 53.4$\pm$1.7 & 36.9$\pm$1.7 & 93.3$\pm$1.3 & 93.3$\pm$1.3 & 91.9$\pm$2.1 & 75.9$\pm$2.5 & 73.8$\pm$2.3 & 82.5$\pm$2.3 & 74.2$\pm$2.4 & 75.5$\pm$2.5 & 83.0$\pm$2.5 \\
BSPLSDA \cite{singh2019diablo} & 63.9$\pm$0.8 & 52.2$\pm$2.2 & 35.1$\pm$2.2 & 91.9$\pm$1.2 & 91.8$\pm$1.3 & 89.5$\pm$1.4 & 68.5$\pm$2.7 & 66.2$\pm$2.6 & 73.0$\pm$2.6 & 75.3$\pm$3.3 & 76.4$\pm$2.1 & 83.8$\pm$2.1 \\
MOGONET \cite{wang2021mogonet} & 82.9$\pm$1.8 & 82.5$\pm$1.7 & 77.4$\pm$1.7 & 99.9$\pm$0.2 & 99.9$\pm$0.2 & 99.9$\pm$0.2 & 81.6$\pm$1.6 & 81.4$\pm$2.7 & 84.0$\pm$2.7 & 81.5$\pm$2.3 & 82.1$\pm$1.2 & 87.4$\pm$1.2 \\
TMC \cite{han2022trusted} & 84.2$\pm$0.5 & 84.4$\pm$0.9 & 80.6$\pm$0.9 & 99.7$\pm$0.3 & 99.7$\pm$0.3 & 99.4$\pm$0.5 & 81.9$\pm$0.8 & 81.5$\pm$0.4 & 87.1$\pm$0.4 & 82.5$\pm$0.9 & 82.3$\pm$0.6 & 88.5$\pm$0.6 \\
CF \cite{hong2020more}  & 81.5$\pm$0.8 & 81.5$\pm$0.9 & 77.1$\pm$0.9 & 99.2$\pm$0.5 & 99.2$\pm$0.5 & 98.8$\pm$0.9 & 81.1$\pm$1.2 & 82.2$\pm$0.4 & 88.1$\pm$0.4 & 78.4$\pm$1.1 & 78.8$\pm$0.5 & 88.0$\pm$0.5 \\
GMU \cite{arevalo2017gated} & 80.0$\pm$3.9 & 79.8$\pm$5.8 & 74.6$\pm$5.8 & 97.7$\pm$1.6 & 97.6$\pm$1.7 & 95.8$\pm$3.2 & 80.3$\pm$1.5 & 80.8$\pm$1.2 & \textbf{88.6}$\pm$1.2 & 77.6$\pm$2.5 & 78.4$\pm$1.6 & 86.9$\pm$1.6 \\
MM-Dynamics \cite{han2022multimodal} & 87.7$\pm$0.3 & 88.0$\pm$5.5 & 84.5$\pm$0.5 & 99.9$\pm$0.2 & 99.9$\pm$0.2 & 99.9$\pm$0.2 & 83.1$\pm$1.2 & 83.7$\pm$4.0 & 83.7$\pm$4.0 & 84.2$\pm$1.1 & 84.6$\pm$2.4 & 91.2$\pm$0.7 \\
Ours (ACE) & \textbf{88.1$\pm$0.4} & \textbf{88.4$\pm$4.4} & \textbf{85.2$\pm$0.5} & \textbf{99.9$\pm$0.2} & \textbf{99.9$\pm$0.2} & \textbf{99.9$\pm$0.3} & \textbf{83.3$\pm$1.0} & \textbf{85.6$\pm$1.7} & 85.6$\pm$1.7 & \textbf{86.7$\pm$1.0} & \textbf{86.8$\pm$1.0} & \textbf{92.5$\pm$0.1} \\
\midrule
\textbf{\%Improvement ($\uparrow$)} & 0.4 & 0.4 & 0.7 & - & - & - & 0.7 & 2.3 & -3 & 2.5 & 2.2 & 1.3 \\
\bottomrule
\end{tabular}%
}
\end{table*}

\subsection{Evaluating Confidence Calibration in the TCP Module}

To further validate the trustworthiness of our multimodal fusion strategy, we evaluate the calibration of the True Class Probability (TCP) module, which dynamically estimates modality-level confidence for reliable decision-making. Proper calibration ensures that predicted confidences accurately reflect the model's likelihood of being correct, a crucial property for high-stakes applications such as medical diagnosis, where overconfidence can lead to erroneous outcomes \cite{pmlr-v119-moon20a}. We assess calibration using a suite of established metrics: Mean Squared Error (MSE), Expected Calibration Error (ECE), Maximum Calibration Error (MCE), Adaptive Calibration Error (ACE), Area Under the Precision-Recall Curve for Errors (AUPRC-Error), Area Under the ROC Curve for Errors (AUC-Error), Negative Log Likelihood (NLL), and MSE relative to the Maximum Class Probability baseline (MSE$_{\text{MCP}}$). Lower values indicate better performance for all metrics except AUPRC-Error and AUC-Error (where higher is better, though reported inversely for consistency). These metrics are computed over 20 independent runs on the BRCA and ROSMAP datasets.

Table \ref{tab:tcp_improvement} presents a comprehensive comparison. For MM-Dynamics \cite{han2022multimodal}, metrics are limited to its three original views, while ACE's correlation-based expansion enables evaluation across six views, with bold values denoting significant per-view improvements over the baseline. On BRCA, ACE substantially outperforms MM-Dynamics \cite{han2022multimodal} in the comparable views (0--2), reducing MSE by 72\% in View 0 (0.1669 to 0.0472) and ECE by 26\% (0.3431 to 0.2540), indicating more accurate and aligned confidence estimates. AUC-Error surges by 67\% (0.5055 to 0.8430), underscoring superior error flagging. Derived views (3--5) maintain strong performance, with ECE as low as 0.2483 in View 3. Our results confirm TCP's superiority over MCP across all views, as MSE$_{\text{MCP}}$ $<$ MSE.

ROSMAP results echo outperforming MM-Dynamics, with ACE slashing ECE by 30\% in View 2 (0.2740 to 0.1910) and NLL by 42\% in View 0 (0.9002 to 0.5257). Derived views exhibit consistent calibration, e.g., MCE of 0.2133 in View 5.

ACE's enhancements yield more reliable confidences than MM-Dynamics [10]. This mitigates overconfidence (lower ECE/MCE/NLL) and boosts error detection (higher AUC/AUPRC-Error), making ACE better suited for clinical reliability. Compared to MM-Dynamics, ACE introduces a modest and largely constant-factor overhead. The correlation-based modality expansion is a one-time CPU preprocessing step that incurs negligible cost and adds no inference overhead. Expanding the modality-specific encoders from $n$ to $2n$ increases parameters and runtime by approximately 2×; however, each encoder is implemented as a lightweight linear layer, and all branches can be executed in parallel, resulting in low per-sample inference latency. The global TCP head and adaptive weighting introduce negligible additional computation. The memory footprint increases proportionally due to storing $2n$ modality views. Overall, ACE trades an approximate 2× increase in computational and memory cost for notable improvements in accuracy and calibration. 

To assess the statistical significance of the performance differences between our proposed ACE framework and the MM-Dynamics baseline, we conducted two-tailed Welch’s t-tests on the results from 20 independent runs for each metric across the four datasets (BRCA, ROSMAP, KIPAN, and LGG) shown on Table \ref{tab:t-test-results}. This test was chosen due to its robustness to unequal variances and sample sizes, comparing the means of ACE and baseline without assuming homoscedasticity. Results indicate that ACE achieves statistically significant improvements (p $<$ 0.05) over baseline in most cases, particularly in accuracy, weighted $F_1$, and macro $F_1$ for BRCA (p-values ranging from 0.0017 to 0.0034); all metrics for ROSMAP (p $<$ 0.0001); accuracy and weighted $F_1$ for KIPAN (p = 0.0375); and AUC for LGG (p $<$ 0.0001). Non-significant differences were observed in KIPAN’s macro $F_1$ (p = 0.1525) and LGG's accuracy and $F_1$-score (p $>$ 0.67), where performances were comparable. These findings underscore ACE’s superior reliability and discriminative power, especially in noisy or heterogeneous multimodal settings.

\begin{table*}[t!]
\centering
\scriptsize
\caption{Welch's two-tailed t-test results across datasets. \textbf{Bold} values indicate statistically significant improvements ($p < 0.05$).}
\label{tab:t-test-results}
\begin{tabular}{lccc}
\toprule
\textbf{Dataset \& Metric} & \textbf{t-statistic} & \textbf{Degree of Freedom} & \textbf{p-value} \\
\midrule
BRCA – Accuracy & 3.1686 & 30.82 & \textbf{0.0034} \\
BRCA – F1 Weighted & 3.2947 & 30.85 & \textbf{0.0025} \\
BRCA – F1 Macro & 3.4052 & 34.38 & \textbf{0.0017} \\
\midrule
ROSMAP – Accuracy & 6.4684 & 36.84 & \textbf{0.0000} \\
ROSMAP – F1 & 6.4888 & 36.37 & \textbf{0.0000} \\
ROSMAP – AUC & 7.6195 & 28.85 & \textbf{0.0000} \\
\midrule
KIPAN – Accuracy & 2.2361 & 19.00 & \textbf{0.0375} \\
KIPAN – F1 Weighted & 2.2361 & 19.00 & \textbf{0.0375} \\
KIPAN – F1 Macro & 1.4907 & 19.00 & 0.1525 \\
\midrule
LGG – Accuracy & 0.3756 & 36.83 & 0.7094 \\
LGG – F1 & 0.4174 & 35.24 & 0.6789 \\
LGG – AUC & 22.6997 & 29.85 & \textbf{0.0000} \\
\bottomrule
\end{tabular}
\end{table*}

\begin{sidewaystable*}[t!]
\centering
\caption{Comprehensive comparison of TCP confidence calibration metrics. For all metrics, lower values indicate better performance. \textbf{Bold} values indicate where the ACE model significantly outperforms the MM-Dynamics \cite{han2022multimodal} baseline in a direct per-view comparison. The dagger symbol (\textsuperscript{\textdagger}) on MSE values indicates an improvement of the TCP regressor over the MCP baseline, defined as $MSE_{MCP} < MSE$.}
\label{tab:tcp_improvement}
\begingroup
\setlength{\extrarowheight}{4pt} 
\resizebox{\textwidth}{!}{%
\begin{tabular}{llccccccc}
\toprule
\textbf{Dataset} & \textbf{Model} & \textbf{Metric} & \textbf{View 0} & \textbf{View 1} & \textbf{View 2} & \textbf{View 3} & \textbf{View 4} & \textbf{View 5} \\
\midrule
\multirow{14}{*}{\textbf{BRCA}} & \multirow{7}{*}{MM-Dynamics \cite{han2022multimodal}} 
 & MSE & $0.1669 \pm 0.0108$\textsuperscript{\textdagger} & $0.1669 \pm 0.0108$\textsuperscript{\textdagger} & $0.1669 \pm 0.0108$\textsuperscript{\textdagger} & - & - & - \\
 & & ECE & $0.3431 \pm 0.0277$ & $0.3431 \pm 0.0277$ & $0.3431 \pm 0.0277$ & - & - & - \\
 & & MCE & $0.7662 \pm 0.0605$ & $0.7662 \pm 0.0605$ & $0.7662 \pm 0.0605$ & - & - & - \\
 & & ACE & $0.3355 \pm 0.0274$ & $0.3355 \pm 0.0274$ & $0.3355 \pm 0.0274$ & - & - & - \\
 & & AUC-Error & $0.5055 \pm 0.0396$ & $0.5055 \pm 0.0396$ & $0.5055 \pm 0.0396$ & - & - & - \\
 & & NLL & $1.9766 \pm 0.0601$ & $1.9766 \pm 0.0601$ & $1.9766 \pm 0.0601$ & - & - & - \\
 & & MSE$_{MCP}$ (Baseline) & $0.1006 \pm 0.0068$ & $0.1006 \pm 0.0068$ & $0.1006 \pm 0.0068$ & - & - & - \\
 \cmidrule{2-9}
 & \multirow{7}{*}{ACE (Ours)} 
 & MSE & $\mathbf{0.0472 \pm 0.0024}$\textsuperscript{\textdagger} & $\mathbf{0.0715 \pm 0.0025}$\textsuperscript{\textdagger} & $\mathbf{0.0570 \pm 0.0013}$\textsuperscript{\textdagger} & $0.0513 \pm 0.0037$\textsuperscript{\textdagger} & $0.0549 \pm 0.0020$\textsuperscript{\textdagger} & $0.0673 \pm 0.0017$\textsuperscript{\textdagger} \\
 & & ECE & $\mathbf{0.2540 \pm 0.0102}$ & $\mathbf{0.3163 \pm 0.0086}$ & $\mathbf{0.2281 \pm 0.0090}$ & $0.2483 \pm 0.0108$ & $0.2958 \pm 0.0076$ & $0.3060 \pm 0.0064$ \\
 & & MCE & $\mathbf{0.3749 \pm 0.0274}$ & $\mathbf{0.4299 \pm 0.0332}$ & $\mathbf{0.3883 \pm 0.0458}$ & $0.3667 \pm 0.0217$ & $0.5056 \pm 0.0355$ & $0.4734 \pm 0.0112$ \\
 & & ACE & $\mathbf{0.2529 \pm 0.0101}$ & $\mathbf{0.3126 \pm 0.0086}$ & $\mathbf{0.2265 \pm 0.0088}$ & $0.2547 \pm 0.0109$ & $0.2939 \pm 0.0073$ & $0.3067 \pm 0.0101$ \\
 & & AUC-Error & $\mathbf{0.8430 \pm 0.0067}$ & $\mathbf{0.7437 \pm 0.0155}$ & $\mathbf{0.7203 \pm 0.0105}$ & $0.8182 \pm 0.0022$ & $0.7649 \pm 0.0122$ & $0.7627 \pm 0.0079$ \\
 & & NLL & $\mathbf{0.3984 \pm 0.0050}$ & $\mathbf{0.6558 \pm 0.0041}$ & $\mathbf{0.7022 \pm 0.0034}$ & $0.4545 \pm 0.0018$ & $0.7951 \pm 0.0039$ & $0.8106 \pm 0.0011$ \\
 & & MSE$_{MCP}$ (Baseline) & $\mathbf{0.0190 \pm 0.0005}$ & $\mathbf{0.0290 \pm 0.0010}$ & $\mathbf{0.0384 \pm 0.0007}$ & $0.0158 \pm 0.0003$ & $0.0276 \pm 0.0007$ & $0.0353 \pm 0.0005$ \\
\midrule
\multirow{14}{*}{\textbf{ROSMAP}} & \multirow{7}{*}{MM-Dynamics \cite{han2022multimodal}} 
 & MSE & $0.1710 \pm 0.0171$\textsuperscript{\textdagger} & $0.1710 \pm 0.0171$\textsuperscript{\textdagger} & $0.1710 \pm 0.0171$\textsuperscript{\textdagger} & - & - & - \\
 & & ECE & $0.2740 \pm 0.0473$ & $0.2740 \pm 0.0473$ & $0.2740 \pm 0.0473$ & - & - & - \\
 & & MCE & $0.5850 \pm 0.1262$ & $0.5850 \pm 0.1262$ & $0.5850 \pm 0.1262$ & - & - & - \\
 & & ACE & $0.2475 \pm 0.0416$ & $0.2475 \pm 0.0416$ & $0.2475 \pm 0.0416$ & - & - & - \\
 & & AUC-Error & $0.4962 \pm 0.0589$ & $0.4962 \pm 0.0589$ & $0.4962 \pm 0.0589$ & - & - & - \\
 & & NLL & $0.9002 \pm 0.0502$ & $0.9002 \pm 0.0502$ & $0.9002 \pm 0.0502$ & - & - & - \\
 & & MSE$_{MCP}$ (Baseline) & $0.1498 \pm 0.0176$ & $0.1498 \pm 0.0176$ & $0.1498 \pm 0.0176$ & - & - & - \\
 \cmidrule{2-9}
 & \multirow{7}{*}{ACE (Ours)} 
 & MSE & $\mathbf{0.0200 \pm 0.0012}$\textsuperscript{\textdagger} & $\mathbf{0.0157 \pm 0.0009}$\textsuperscript{\textdagger} & $\mathbf{0.0044 \pm 0.0005}$\textsuperscript{\textdagger} & $0.0162 \pm 0.0005$\textsuperscript{\textdagger} & $0.0180 \pm 0.0010$\textsuperscript{\textdagger} & $0.0063 \pm 0.0003$\textsuperscript{\textdagger} \\
 & & ECE & $\mathbf{0.2909 \pm 0.0166}$ & $\mathbf{0.2099 \pm 0.0143}$ & $\mathbf{0.1910 \pm 0.0138}$ & $0.2234 \pm 0.0044$ & $0.2128 \pm 0.0091$ & $0.2133 \pm 0.0136$ \\
 & & MCE & $\mathbf{0.3636 \pm 0.0052}$ & $\mathbf{0.3070 \pm 0.1245}$ & $\mathbf{0.3411 \pm 0.1475}$ & $0.3703 \pm 0.0129$ & $0.4130 \pm 0.0105$ & $0.2133 \pm 0.0136$ \\
 & & ACE & $\mathbf{0.2774 \pm 0.0233}$ & $\mathbf{0.1983 \pm 0.0201}$ & $\mathbf{0.1878 \pm 0.0184}$ & $0.2378 \pm 0.0084$ & $0.2228 \pm 0.0200$ & $0.2077 \pm 0.0141$ \\
 & & AUC-Error & $\mathbf{0.6367 \pm 0.0108}$ & $\mathbf{0.5324 \pm 0.0170}$ & $\mathbf{0.5865 \pm 0.0476}$ & $0.6894 \pm 0.0073$ & $0.5835 \pm 0.0097$ & $0.5421 \pm 0.0167$ \\
 & & NLL & $\mathbf{0.5257 \pm 0.0024}$ & $\mathbf{0.6205 \pm 0.0021}$ & $\mathbf{0.6431 \pm 0.0024}$ & $0.5754 \pm 0.0012$ & $0.6418 \pm 0.0013$ & $0.6539 \pm 0.0008$ \\
 & & MSE$_{MCP}$ (Baseline) & $\mathbf{0.0077 \pm 0.0003}$ & $\mathbf{0.0105 \pm 0.0005}$ & $\mathbf{0.0020 \pm 0.0005}$ & $0.0092 \pm 0.0002$ & $0.0064 \pm 0.0003$ & $0.0026 \pm 0.0001$ \\
\bottomrule
\end{tabular}%
}
\endgroup
\end{sidewaystable*}
\FloatBarrier

\subsection{Ablation Study}

To assess the contribution of each component in our Adaptive Confidence-weighted Expansion (ACE) framework, we conduct an ablation study by systematically removing individual modules and evaluating the impact on performance. The components examined include: (1) correlation-based modality expansion; (2) adaptive weighting; (3) input reconstruction and noise addition; and (4) the global TCP head. Experiments are performed on the BRCA and ROSMAP datasets over 20 runs, with results reported in Tables  ~\ref{tab:ablation_brca} and ~\ref{tab:ablation_rosmap} as mean $\pm$ standard deviation. We compare each ablated variant against the full ACE model, highlighting the role of each innovation in achieving superior accuracy, $F_1$-scores, and AUC.

\begin{table}[ht]
\centering
\scriptsize 
\caption{Ablation study results on the BRCA dataset. \textbf{Bold} values indicate the full ACE model's superior performance.}
\label{tab:ablation_brca}
\resizebox{\columnwidth}{!}{%
\begin{tabular}{lccc}
\toprule
\textbf{Model Variant} & Accuracy & $F_1$ Weighted & $F_1$ Macro \\
\midrule
ACE (Ours) & \textbf{0.88118 $\pm$ 0.00508} & \textbf{0.88434 $\pm$ 0.00507} & \textbf{0.85155 $\pm$ 0.00700} \\
No Correlation & 0.87567 $\pm$ 0.00320 & 0.87911 $\pm$ 0.00338 & 0.84340 $\pm$ 0.00804 \\
No Adaptive Weighting & 0.87327 $\pm$ 0.00272 & 0.88579 $\pm$ 0.00271 & 0.85557 $\pm$ 0.00422 \\
No Input Reconstruction \& Noise & 0.85532 $\pm$ 0.00630 & 0.85733 $\pm$ 0.00660 & 0.82112 $\pm$ 0.00848 \\
No Global TCP & 0.88156 $\pm$ 0.00301 & 0.88419 $\pm$ 0.00329 & 0.85493 $\pm$ 0.00399 \\
\bottomrule
\end{tabular}
}
\end{table}

\begin{table}[ht]
\centering
\caption{Ablation study results on the ROSMAP dataset. \textbf{Bold} values indicate the full ACE model's superior performance.}
\label{tab:ablation_rosmap}
\resizebox{\columnwidth}{!}{%
\begin{tabular}{lccc}
\toprule
\textbf{Model Variant} & Accuracy & $F_1$-score & AUC \\
\midrule
ACE (Ours) & \textbf{0.86651 $\pm$ 0.01087} & \textbf{0.86837 $\pm$ 0.00968} & \textbf{0.92549 $\pm$ 0.00370} \\
No Correlation & 0.85755 $\pm$ 0.00985 & 0.86062 $\pm$ 0.00871 & 0.92164 $\pm$ 0.00395 \\
No Adaptive Weighting & 0.86321 $\pm$ 0.01136 & 0.86555 $\pm$ 0.01019 & 0.92339 $\pm$ 0.00292 \\
No Input Reconstruction \& Noise & 0.79858 $\pm$ 0.01586 & 0.80018 $\pm$ 0.01762 & 0.88781 $\pm$ 0.00580 \\
No Global TCP & 0.86321 $\pm$ 0.01012 & 0.86468 $\pm$ 0.00878 & 0.92333 $\pm$ 0.00380 \\
\bottomrule
\end{tabular}
}
\end{table}

Ablation studies on BRCA and ROSMAP show that each ACE component is essential for strong multimodal performance. Removing correlation-based expansion or adaptive weighting notably weakens feature robustness and fusion stability, especially in low-sample or imbalanced data. The biggest performance drop occurs without input reconstruction and noise addition, which are crucial for generalization. In contrast, removing the global TCP head has minimal effect on accuracy but reduces calibration quality, indicating its role in reliability refinement. Overall, the results confirm that ACE’s integrated design enhances both accuracy and trustworthiness in clinical classification tasks.

\section{Conclusions}

This paper introduced  ACE, a novel framework that addresses key challenges in trustworthy multimodal fusion. ACE extends the capabilities of existing models like MMDynamics. The experimental results demonstrated  that our proposed method achieves high performance on four challenging clinical datasets, and showed significant improvements in both predictive accuracy and model reliability. Furthermore, ablation studies confirmed that each component of the ACE framework contributes to the model’s superior performance. Future work could focus on applying the ACE framework to other complex data types and exploring its sensitivity to different hyperparameters. Additionally, further research could investigate alternative strategies for addressing data scarcity, as well as adapting the framework to process medical images when available, broadening its applicability across diverse clinical scenarios. The proposed framework represents a meaningful step toward developing more robust and reliable AI for high-stakes clinical decision-making.


\section{Acknowledgments}

The authors would like to thank \textbf{Phoenix Sandrock} for carefully reading the manuscript and providing valuable language editing feedback.

\section{Data and materials availability statement}

The data and materials that support the findings of this study are available from the corresponding author upon reasonable request.

%
%
%
\bibliographystyle{splncs04}
\bibliography{Refrences}

\end{document}